\documentclass[runningheads]{llncs}

\usepackage{graphicx}
\usepackage{amsmath,amssymb} 
\usepackage{color}

\usepackage{hyperref}
\usepackage{marvosym}

\begin{document}
\pagestyle{headings}
\mainmatter

\def\ACCV22SubNumber{505}  

\title{FaceFormer: Scale-aware Blind Face Restoration with Transformers} 

\author{Aijin Li$^1$, \href{mailto:leegeun@yonsei.ac.kr}{Gen Li$^1$\thanks{Corresponding author}}, 
Lei Sun$^1$, Xintao Wang$^2$}
\institute{Platform Technologies, Tencent PCG $^1$  \protect\\ ARC Lab, Tencent PCG $^2$}

\maketitle

\begin{abstract}
Blind face restoration usually encounters with diverse scale face inputs, especially in the real world. However, most of the current works support specific scale faces, which limits its application ability in real-world scenarios. 
In this work, we propose a novel scale-aware blind face restoration framework, named {\bf FaceFormer}, which formulates facial feature restoration as scale-aware transformation. The proposed Facial Feature Up-sampling (FFUP) module dynamically generates upsampling filters based on the original scale-factor priors, which facilitate our network to adapt to arbitrary face scales. Moreover, we further propose the facial feature embedding (FFE) module which leverages transformer to hierarchically extract diversity and robustness of facial latent. Thus, our FaceFormer achieves fidelity and robustness restored faces, which possess realistic and symmetrical details of facial components.
Extensive experiments demonstrate that our proposed method trained with synthetic dataset generalizes better to a natural low quality images than current state-of-the-arts.
\end{abstract}

\section{Introduction}

Blind face restoration can be defined as the ability to recovering the high quality (HQ) realistic details from its low quality (LQ) ones without knowing the degradation types, such as low-resolution, blur, noisy and lossy compression. In many real-world scenarios, the restoration performance of the face images is degraded by variant posture, occlusion, cluttered background, etc.. 

To handle the aforementioned issues, various deep learning-based approaches have been proposed and have shown impressive performance in face restoration. These approaches can be divided into three categories, such as facial prior guided, facial attribute information and identity information \cite{wang2020deep}. 
Unlike general image restoration,  the face prior guided approaches utilize extra facial prior knowledge to recover facial details such as wrinkles, pupils, and eyelashes, etc.. Hence, many related literature adopts facial parsing maps, landmarks and heatmaps. However, these prior information are usually difficult to correctly extract from LQ image due to severe degradation. 
In addition, the facial attribute-based approaches utilize a single or multiple semantic information, such as gender, hair color, age, expression and mustache, etc.. The restored face images bring in different artifacts due to one-to-many maps from LQ images to HQ ones. 
For the identity information, those methods are always used for keeping the identity consistency between its restorations and ground truths. However, those methods are confined to the capacity of face dictionaries or high quality references. 

To our best knowledge, although recent approaches have achieved promising performance, most the face restoration approaches are developed for better super-resolving under a specific resolution of low-quality faces. To guarantee the specific input resolution, conventional upsampling operations, such as bicubic interpolation \cite{wang2021towards,chen2021progressive}, are usually employed. However, the interpolation-based upsamplers enlarge the image resolution only based on own image signals, without bringing any more information. Instead, they often introduce some side effects, such as artifacts, noise amplification and blurring restorations \cite{wang2020deep}. 

Therefore, to overcome these drawbacks, this work proposes a novel scale-aware blind face restoration framework, named {\bf FaceFormer}. Our method consists of facial feature up-sampling, facial feature embedding and facial feature generator. 

Specifically, in this work, we first integrate a super-resolving architecutre with a fractional sampler module to refine more accurate scale-aware features, which makes the proposed method applicable and flexible in the arbitray scale factor for the blind face restoration. After that, inspired by recent success of transformer \cite{liu2021swin}, we leverage the hierarchical transformer blocks for the facial feature embedding with some different non-overlapping local windows and cross-window connection to extract diversity and robustness of facial latent feature. Moreover, the generated face images by proposed method usually possess a realistic and symmetrical facial details, such as pupil type, eyebrow and eyelash etc.. 

We summarize the our contributions as follows.

$\left(1\right)$ We propose a novel fractional up-sampler for blind face restoration. The module manages to generate scale-aware super-resolutions of low quality faces in the real-world scenarios. 

$\left(2\right)$ Furthermore, we propose the FaceFormer framework with the facial feature up-sampling module and facial feature embedding module to improve fidelity and robustness. The facial feature embedding module utilizes some different non-overlapping local windows and cross-window connection to extract the more robust and diverse tiny facial component features.

$\left(3\right)$ Extensive experiments demonstrate that our proposed method trained with synthetic datasets generalizes better to natural low quality images than current state-of-the-arts.

\begin{figure}[t]
	\centering
	\includegraphics[width=11cm, height=7cm]{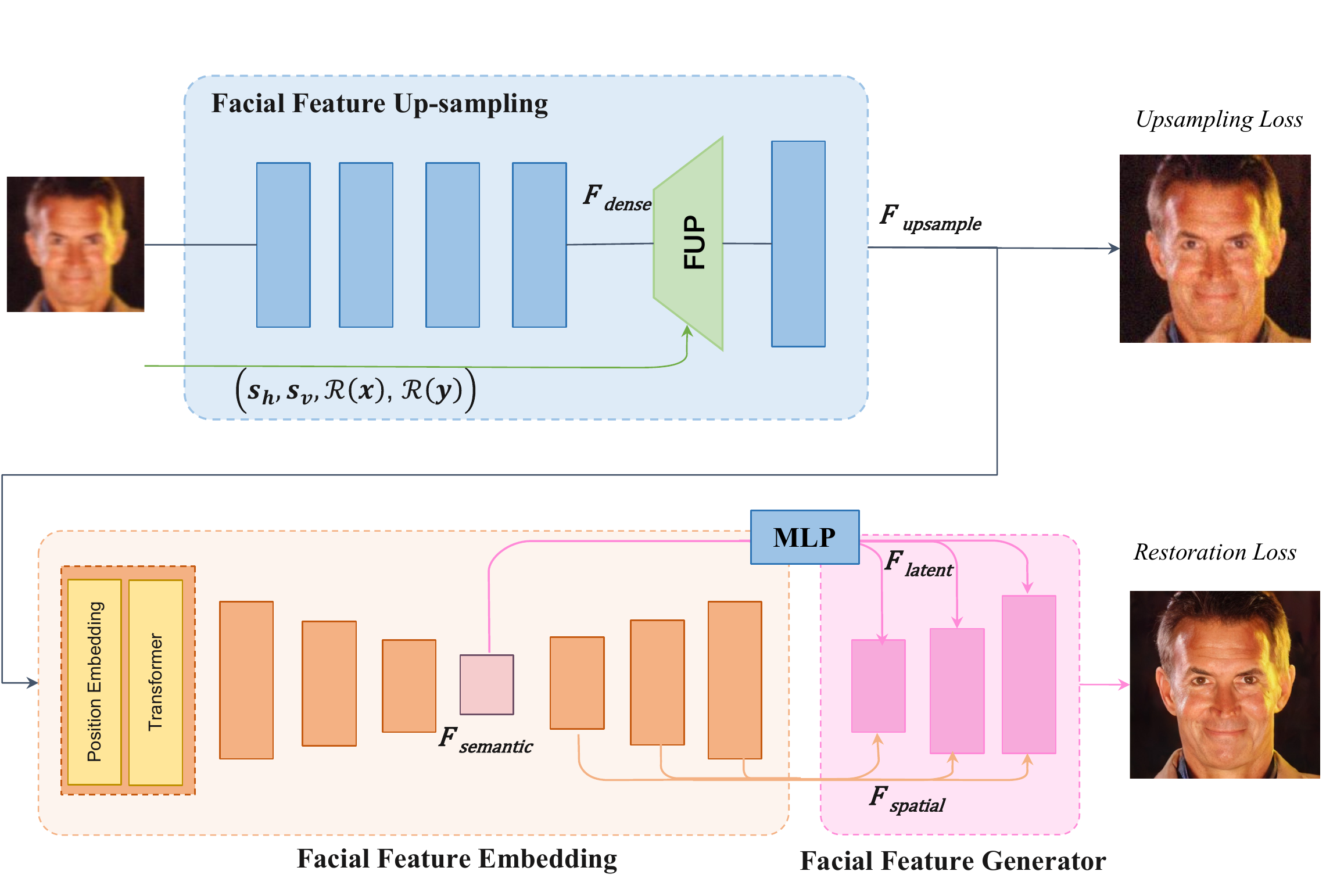}
	\caption{The framework of proposed FaceFormer. It is mainly composed of Facial Feature Up-sampling, Facial Feature Embedding and and Facial Feature Generator.}
	\label{fig:FaceFormer}
\end{figure}

\section{Related Works}

\subsection{Face Super-Resolution}

Face super resolution is often studied as a basic task for face restoration. Research \cite{zhu2016deep} proposed a bidirection-framework that alternately optimizes facial restoration and dense corresponding field estimation in a cascade manner. Research \cite{yu2016ultra} employed generative adversarial networks (GAN) to directly super-resolve low resolution (LR) inputs. More researchers further exploited challenging problems like unaligned faces \cite{yu2017face}, noisy faces \cite{yu2017hallucinating} and faces with different attributes \cite{yu2018super}. Wavelet-SRNet \cite{huang2017wavelet} manages to generate high resolution (HR) wavelet coefficient sequence for HR image reconstruction instead of directly inferring HR image. Recent researches utilize additional face priors to improve performance. 
FSRNet \cite{chen2018fsrnet} employs face landmark heatmaps and parsing maps to refine the super-resolution results, especially for misaligned LR faces. 
Instead, SuperFAN \cite{bulat2018super} forms face SR and landmark prediction as multi-task learning. 
In addition to the LR super-resolution branch, research \cite{yu2018face} extra exploites a face component heat map to keep the face structure. 
Research \cite{kim2019progressive} employs the difference heatmap as weights for multi-scale facial attention loss to better landmark restorations. 
Nonetheless, most of these researches succeed a general encoder-decoder framework without full application of face priors. It is not well suitable to handle real-world scenarios.


\subsection{Blind Face Restoration}

To deal with blind face restoration, recent works typically tried to adopted face-specific information, geometry priors, facial attribute information and identity information. The geometry priors include facial landmarks \cite{chen2018fsrnet,kim2019progressive,zhu2016deep}, face parsing maps \cite{shen2018deep,chen2021progressive,chen2018fsrnet} and facial component heatmaps \cite{yu2018face}. 
However, these prior information are usually tough to correctly extract from LQ image due to severe degradation in real-world scenarios.

For attribute information, it is accessible without LQ face images, which is distinct from face structure prior information. AGCycleGAN \cite{lu2018attribute}, FSRSA variants \cite{yu2018super} and ATNet \cite{li2020learning} versions are directly concatenate attribute information and LR face image. The aforementioned methods take no consideration for the absence of attributes, limiting their applicability in real world scenes. RAAN \cite{xin2019residual} constructs multi-branch residual attention to further aggregate attribute information. Instead, FACN \cite{xin2020facial} proposes a capsule generation block for attribute integration.

Identity information usually refers to HQ face images of the same identity with LR faces. GFRNet \cite{li2018learning} employes single HQ face images to share the same identity. while the variable facial expressions and poses may suppress the restoration of face images. ASFFNet \cite{li2020enhanced} is the foregoer to explore multi-face guided FSR, which selects the best reference image by guidance selection module. Its variants \cite{huang2017arbitrary} further cope with misalignment and illumination differences of multi-face guided face restoration. According to the observation that different people may have similar facial components, research \cite{li2019recovering} builds a component dictionary to advance face restoration. DFDNet \cite{li2020blind} builds multi-scale component dictionaries based on features of the entire dataset.

\begin{figure}[t]
	\centering
	\includegraphics[width=11cm, height=7cm]{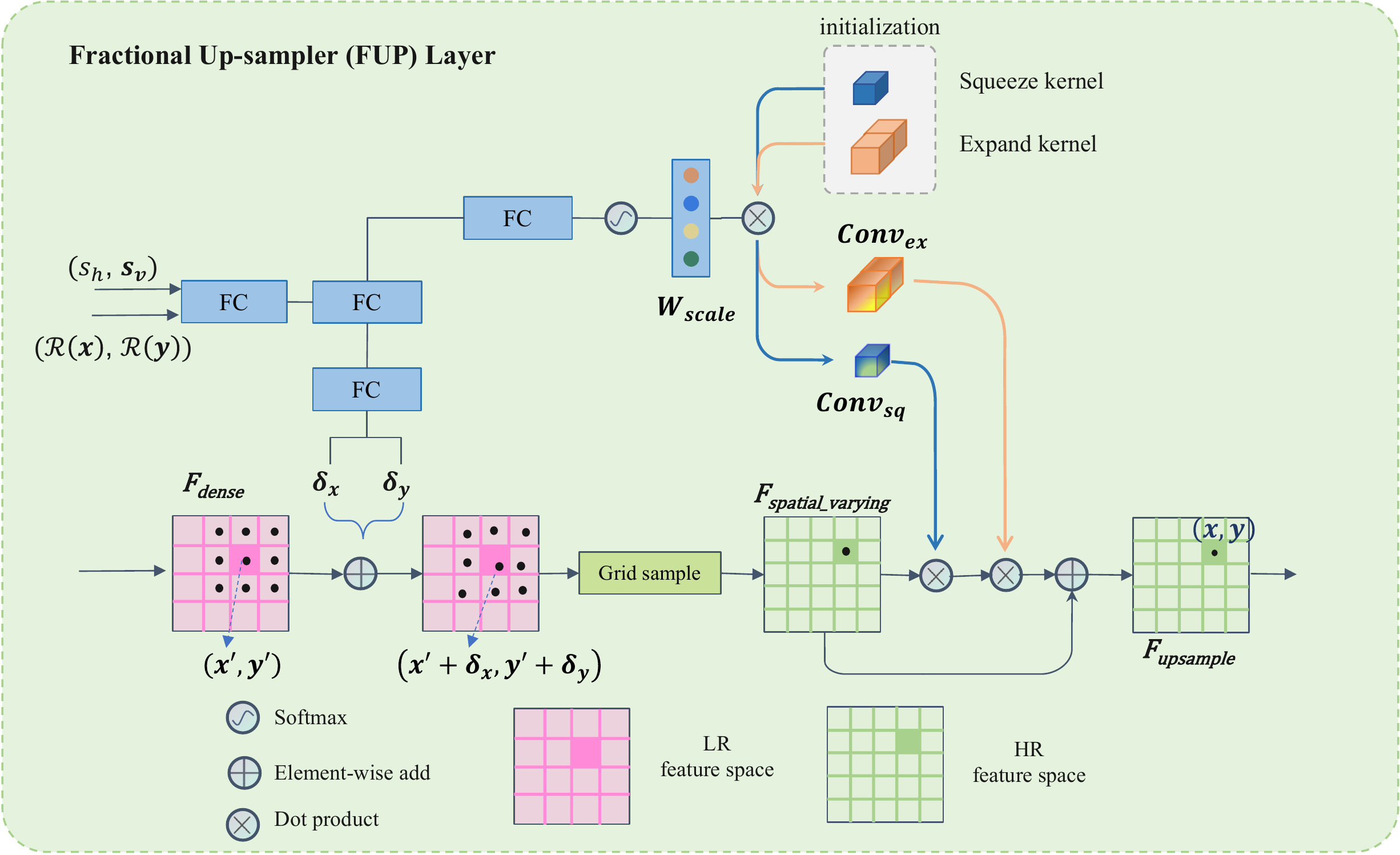}
	\caption{The architecture of Facial Feature Up-sampling Module.}
	\label{fig:FFUP}
\end{figure}

\subsection{Vision Transformer} Recently, natural language processing model Transformer \cite{vaswani2017attention} received wide attention in the computer vision field. It learns to pay attention to important image regions by exploring the global interactions between different regions, and is widely available for image classification \cite{liu2021swin}, object detection \cite{carion2020end}, segmentation \cite{zheng2021rethinking}, crowd counting \cite{liang2021transcrowd} and so on. Thanks to its impressive performance, Transformer is also employed for image restoration \cite{chen2021pre,cao2021video}. Research \cite{chen2021pre} builds a backbone model IPT for various restoration problems based on the standard Transformer. However, IPT works on the condition that there are mass datasets for multi-task learning. Research \cite{cao2021video} applies self-attention mechanism VSR-Transformer for feature fusion in video SR, while image features are still extracted from CNN. In addition, the patch-wise attention in IPT and VSR-Transformer may not well suitable for the whole image restoration.

\section{Methodology}


\begin{figure}[t]
	\centering
	\includegraphics[width=6cm, height=4cm]{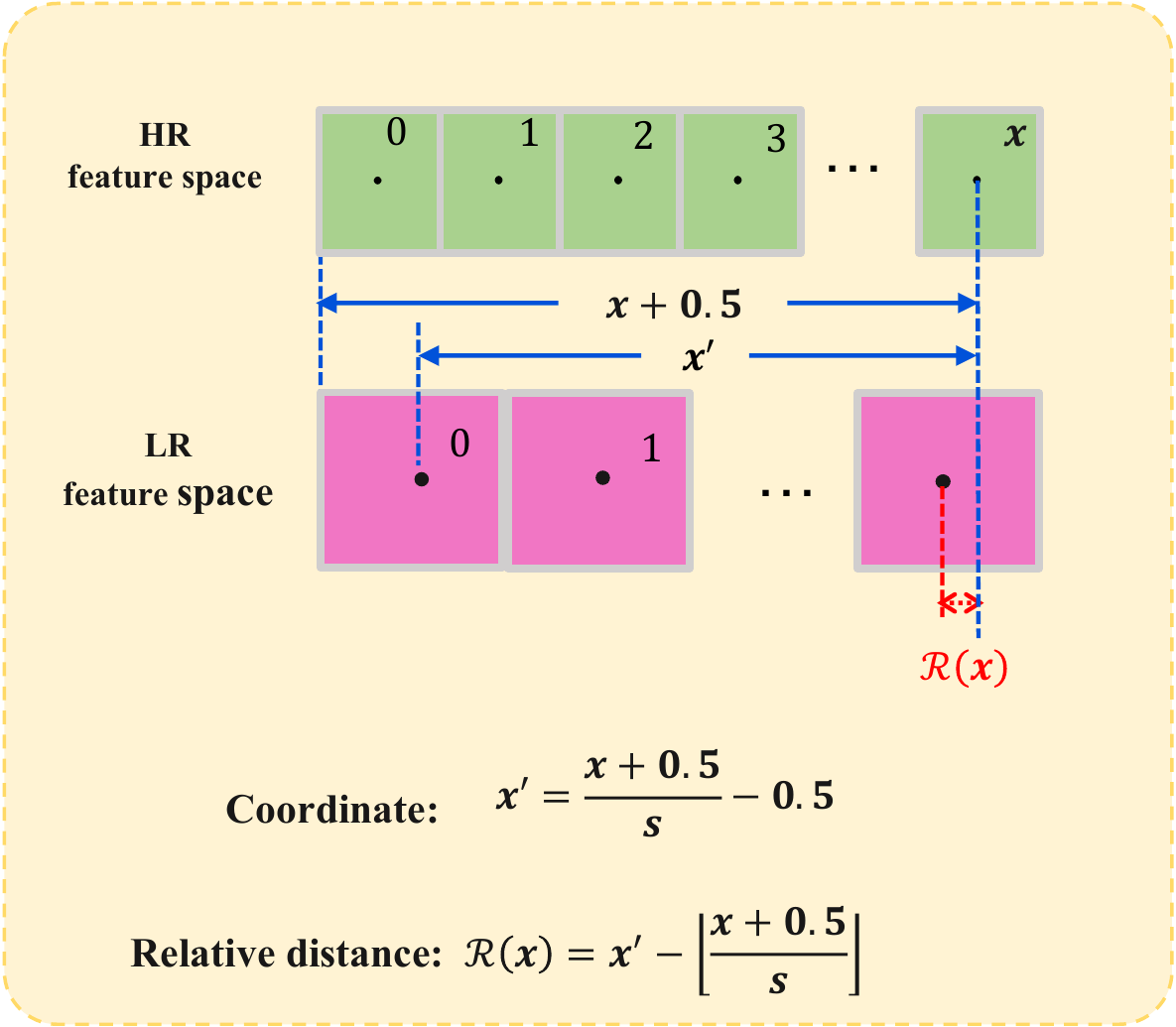}
	\caption{The computation of coordinate ($x^{\prime}$, $y^{\prime}$) and relative distances ($R(x)$ and $R(y)$) in LR feature space}
	\label{fig:scale_info}
\end{figure}

\subsection{Overview of Our FaceFormer}

This section describes the proposed FaceFormer architecture. Given an input face image $I$ suffering from unknown degradation, the aim of blind face restoration is to estimate a high-quality image $\hat{y}$, which is as similar as possible to the ground-truth image $y$ in terms of both realistic and fidelity. 

The architecture of the proposed FaceFormer is shown in Fig.~\ref{fig:FaceFormer}. The FaceFormer is mainly composed of facial feature up-sampling (FFUP), facial feature embedding (FFE) and facial feature generator (such as pretrained Style-GAN2 \cite{karras2020analyzing}). Sequentially, to extract scale-aware upsampled facial features, the arbitrary scale facial image $I$ is firstly fed into the FFUP module with scale factor ($s_{h}$, $s_{v}$) and relative distance ($R(x)$, $R(y)$ for scale-aware feature up-sampling.
Then, the FFE module is proposed to extracted facial semantic features $F_{semantic}$ which removing complicated and severer degradations of the output of FFUP module. The FFE module leverages transformer module that captures long-distance dependencies of the face images, such as symmetrical details of components.  

Finally, the facial feature generator (FFG) module is employed to generate diverse and rich facial details. The facial semantic feature $F_{semantic}$ is mapped to the closest latent features to the generative priors, which is provided by the pre-trained GAN. The latent features contain two kinds informative features:
latent features $F_{latent}$ and spatial features $F_{spatial}$.


To summarise, the proposed FaceFormer achieves realistic results while preserving high fidelity in a coarse-to-fine manner.

\subsection{Facial Feature Up-sampling}

In the real world, face restoration explicitly deals with arbitrary resolution LQ images. Many attempts \cite{wang2021towards,chen2021progressive} employ interpolation-based upsamplings to handle multiple scale factors without considering the scale information during feature learning. Intuitively, since the degradation is directly related to various scale factors, scale information can further be adapted to learn discriminative features to improve HQ image performance.
Moreover, these upsampling methods often introduce some side effects, such as artifacts, noise amplification, blurring restorations \cite{aganj2012removing}. To reduce these effects, we propose a novel scale-aware facial feature up-sampling module instead of interpolation upsamplings of the single image super resolution (SISR). The proposed module dynamically generates upsampling filters conditioned on scale information. The proposed architecture is depicted in Fig.~\ref{fig:FFUP}.

To extract dense facial features $F_{dense}$, given a LQ facial image $I$, then, $I$ is fed into the backbone module of a classical super resolution architecture without any upsamplings and reconstructions, such as SRResNet \cite{ledig2017photo}, and ESRGAN \cite{wang2018esrgan}, etc.. Then, following the $F_{dense}$ features, a scale-aware fractional up-sampling layer is designed for arbitrary-scale upsampling, instead of subpixels \cite{shi2016real} or deconvolution layer. Besides, the proposed upsampling layer can be applied for a fractional super-resolving through a simple yet effective modification. This upsampling layer consists of two stages, spatial-sampling stage and channel-varying stage with fractional scale factor inputs.

Given a pixel $\left(x, y\right)$ in the high resolution (HR) feature space, the pixel is firstly projected into the low resolution (LR) feature space. As shown in Fig.~\ref{fig:scale_info}, its corresponding LR coordinate ($x^{\prime}$, $y^{\prime}$) and relative distances ($R(x)$ and $R(y)$) are calculated as follows:

\begin{align}\left\{\begin{aligned}
\label{equ:LR corrdinate}
	x^{\prime}&=\frac{x+0.5}{s_{h}}-0.5 \\
	y^{\prime}&=\frac{y+0.5}{s_{v}}-0.5 
\end{aligned}\right.\end{align}
\begin{align}\left\{\begin{aligned}
	R(x)&=x^{\prime} -  \left \lfloor\frac{x+0.5}{s_{h}} \right \rfloor \\
	R(y)&=y^{\prime} -  \left \lfloor\frac{y+0.5}{s_{v}} \right \rfloor
\end{aligned}\right.\end{align}

where the $s_{h}$ and $s_{v}$ denote horizontal and vertical scale factor, respectively. For a pixel $\left(x, y\right)$, its corresponding LR coordinate $\left(x^{\prime}, y^{\prime}\right)$ are theoretically projected as equation (\ref{equ:LR corrdinate}). However, the generated coordinate $\left(x^{\prime}, y^{\prime}\right)$ is non integer in most cases. Thus, the floor operation $\lfloor \cdot \rfloor$ are often employed for coordinate quantization. The relative distances ($R(x)$ and $R(y)$) denotes the projection bias from HR feature space to LR feature space.

As shown in Fig.~\ref{fig:FFUP}, the relative distances ($R(x)$, $R(y)$) and horizontal and vertical scale factors ($s_{h}$, $s_{v}$) are firstly concatenated and fed into two Fully Connection (FC) layers for scale-aware feature extraction. The relative distances ($R(x)$, $R(y)$) reflect the information interaction between HR and LR feature space for more sufficient scale-aware feature extraction.
Then, offset filter and scale-aware filter are adopt to the extracted feature for offsets ($\delta_{x}$, $\delta_{y}$) and scale-varying weights $W_{scale}$, respectively.
Here, to reduce the complexity and improve the generalization ability, these two kernels refer to two FC layers as follows:

\begin{align}
    (\delta_{x}, \delta_{y}), W_{scale}=FCs(s_{h}||s_{v}||R(x)||R(y))
\end{align}
where the $||$ denotes concatenation.

Subsequently, the scale-varying weights serve as scale guidance to combine squeeze and expand kernels, bringing about a pair of squeeze $Conv_{sq}$ and expansion $Conv_{ex}$ convolution layers. The weights of these convolution layers are adaptively learned conditioned on scale information, while the interpolation-based upsamplings, such as bicubic upsampling adopts a fixed weight. 


In the spatially-sampling stage, given the predicted horizontal and vertical offsets $\delta_{x}$ and $\delta_{y}$, a $k \times k$ neighborhood centered at $\left( x^{\prime} + \delta_{x}, y^{\prime} + \delta_{y} \right)$ in the previous dense facial feature $F_{dense}$ are upsampled using Grid Sample \cite{wang2021learning} as follows:

\begin{align}
F_{spatial\_varying}=GridSample\left( x^{\prime} + \delta_{x}, y^{\prime} + \delta_{y} \right)_{k \times k}
\end{align}

in the channel-varying stage, the spatially-sampled feature $F_{spatial\_varying}$ is fed into with the scaled channel-varying filters for scale information interaction in channels. The output are fused with $F_{spatial\_varying}$ to generate $F_{upsample}$ in a residual way as follows:

\begin{align}
F_{upsample}=Conv_{ex}(Conv_{sq}(F_{spatial\_varying}))+F_{spatial\_varying}
\end{align}

In general, a conventional up-sampling operator adopts its adjacent pixel filling algorithm, which mainly considers a spatially-wise correlation and does not well consider a channel-wise correlation. However, the proposed FFUP module takes care of the both spatially-wise and channel-wise correlation via a proposed fractional upsampler, which significantly improve performance of arbitrary scale face restoration.

\subsection{Facial Feature Embedding} 

Typically, real-world blind face restoration faces suffer from complicated and severer degraded variations in resolution, blur, noise and JPEG artifacts. The Facial Feature Embedding module is designed to explicitly strengthen aforementioned issues and extracted the facial semantic features $F_{semantic}$, providing reliable information for subsequent modules. 

As shown in Fig.~\ref{fig:FaceFormer}, we first extract facial attribute feature maps $F_{attribute}$ from FFUP module upsampled features $F_{upsample}$ as:
\begin{align}
F_{semantic}=Transformer(F_{upsample})
\end{align}

where $Transformer\left ( \cdot \right )$ is a transformer-based feature extraction module (such as swin transfomer, vision transformer etc.) and it contains a position linear embedding and $K$ residual Swin Transformer blocks (STB). The multi-resolution spatial features $F_{spatial}$ are used to modulate the GAN prior features. Note that we use the last convolution layer follow by the STB. Using a convolution layer at the end of feature extraction can bring the inductive bias of the convolution operation into the Transformer-based network, and lay a better foundation for the later aggregation of shallow and deep features. 



\subsection{Facial Feature Generator}

The pre-trained GAN models capture the facial feature distribution, named generative priors, from its learned weight of convolutions. In this work, following GFP-GAN, we adopt the facial semantic feature from proposed FFE module and then map these feature to the pre-trained styleGAN2 to generate realistic and diverse face texture details. Specifically, given the facial semantic feature $F_{semantic}$, the latent feature $F_{latent}$ are generated as:

\begin{align}
	F_{latent}&=MLP(F_{semantic})
\end{align}

where $MLP\left(\cdot \right)$ denotes multi-layer perceptron layers. The $F_{latent}$ then pass through each convolution layer in the pre-trained GAN (such as styleGAN2) and generate GAN features for each resolution scale.  

\subsection{Model Objectives}
The learning objective of training our FaceFormer consists of: 1) upsampling loss $L_{up}$ that constraints the FFUP outputs $y_{up}$ of arbitrary scale inputs close to the ground-truth $y$. 2) restoration loss $L_{restoration}$ for restored output $\hat{y}$ close to $y$.

\subsubsection{Upsampling Loss} We employ the widely-used L1 loss to further enhance the scale-aware significant face upsampling. The upsampling loss $L_{up}$ are defined as follows:

\begin{align}
L_{up}=\lambda_{1} \left \| y_{up}-y\right \|_{1}
\end{align}

\subsubsection{Restoration Loss} Restoration loss $L_{rest}$ are adopt to constraint the outputs $\hat{y}$ close to the ground-truth $y$. Typically, following the GFP-GAN \cite{wang2021towards}, it contains reconstruction loss $L_{rec}$, adversarial loss $L_{adv}$, component loss $L_{comp}$ and identity loss $L_{id}$ as follows:

\begin{align}
L_{rest}=\lambda_{rec} L_{rec}+\lambda_{adv} L_{adv}+\lambda_{comp} L_{comp}+\lambda_{id} L_{id}
\end{align}

The restoration loss hyper-parameters are set as: $\lambda_{rec}=0.1, \lambda_{adv} =0.1, \lambda_{comp}=0.1, \lambda_{id}=1$.

The overall model objective is a combination of the above losses:
\begin{align}
L_{total}=\lambda_{1} L_{up}+\lambda_{2} L_{rest}
\end{align}

The total loss hyper-parameters $\left\{\lambda_{1}, \lambda_{2}\right\}$ are set as $\left\{0.01, 0.1\right\}$, respectively.

\section{Experiments}
\subsection{Datasets and Implementation}
\subsubsection{Training Datasets.}  We adapt the FFHQ \cite{karras2019style} as our training set, which consists of 70,000 high-quality images with size of $1024\times1024$. We resize the images to $512\times512$ with bilinear downsampling as the ground-truth high quality images. 

The proposed FaceFormer is trained on the synthetic data under the degradation model:
\begin{align}
		I=\left [ \left (y\bigotimes K_{\sigma}\right )\downarrow_{r} + n_{\delta}\right ]_{JPEG_{q}}
\end{align}

Given high quality image $y$,  Gaussian blur kernel $ K_{\sigma}$ is firstly employed to convolve with $y$ under parameter $ \sigma$. Then the convolved image is downsampled with a scale factor $r$. Subsequently, the resulting image suffers from additive white Gaussian noise $n_{\delta}$ and finally it is compressed by JPEG with quality factor $q$. Following the practice in GFP-GAN \cite{wang2021towards}, we randomly sample the parameters $\sigma$, $\delta$, $r$ and $q$ from $ \lbrace{0.2:10\rbrace}$, $\lbrace{1 : 8\rbrace}$, $\lbrace{0 : 15\rbrace}$ and $\lbrace{60 : 100\rbrace}$,  respectively. The color jittering is also added during training for color enhancement.
\subsubsection{Testing Datasets.} The FaceFormer is tested on one synthetic dataset and three different real-world datasets. All these datasets have no overlap with our training dataset. The following provide a brief introduction here. 

\textbf{CelebA-Test} is the synthetic dataset with 3,000 CelebA-HQ images from its testing partition \cite{liu2015deep}. The generation is the same way as that during training.

\textbf{LFW-Test.} LFW \cite{huang2008labeled} involves low-quality images in the wild. Here, the first image of each identity in the validation partition is assembled as test set, which contains 1711 images.

\textbf{CelebChild-Test} \cite{wang2021towards} contains 180 child faces of celebrities collected from the Internet. They are low-quality and many of them are black-and-white old photos.

\textbf{WebPhoto-Test} \cite{wang2021towards} is consists of 407 facial images cropped from 188 low-quality photos in real life. These photos have diverse and complicated degradation. Some of them are old photos with very severe degradation on both details and color.
\subsubsection{Implementation.}  The Transformer for degradation removal module consists of encoder and decoder. The encoder contains four downsamples with $\lbrace{2, 4, 6, 2 \rbrace}$ Swin Transformer \cite{liu2021swin} blocks, respectively. And the decoder contains seven upsamples, each with a residual block. Following the work \cite{wang2021towards}, we also adopt the pretrained StyleGAN2 \cite{karras2020analyzing} with $512 \times 512$ outputs as our generative facial prior. 

\setlength{\tabcolsep}{4pt}
\begin{table}
	\begin{center}
		\caption{Quantitative comparison on CelebA-Test for blind face restoration. \textcolor{red}{Red} and \textcolor{blue}{blue} indicates the best and the second best performance. $(*)$ denotes finetuning on our training set.}
		\label{table:celeba_comparisons}
		\begin{tabular}{l|lll|ll}
			\hline
			Methods          & LPIPS $\downarrow$  & FID $\downarrow$   & NIQE $\downarrow$  & PSNR  $\uparrow$ & SSIM $\uparrow$   \\ \hline
			Input            & 0.4866 & 143.98 & 13.440 & 25.35 & \textcolor{blue}{0.6848} \\
			DeblurGANv2$(*)$ \cite{kupyn2019deblurgan}     & 0.4001 & 52.69  & 4.917  & \textcolor{blue}{25.91} & 0.6952 \\
			Wan et al.  \cite{wan2020bringing}      & 0.4826 & 67.58  & 5.356  & 24.71 & 0.6320 \\
			HiFaceGAN  \cite{yang2020hifacegan}       & 0.4770 & 66.09  & 4.916  & 24.92 & 0.6195 \\
			DFDNet \cite{li2020blind}          & 0.4341 & 59.08  & 4.341  & 23.68 & 0.6622 \\
			PSFRGAN \cite{chen2021progressive}     & 0.4240 & 47.59  & 5.123  & 24.71 & 0.6557 \\ \hline
			mGANprior \cite{gu2020image}       & 0.4584 & 82.27  & 6.422  & 24.30 & 0.6758 \\
			PULSE \cite{menon2020pulse}           & 0.4851 & 67.56  & 5.305  & 21.61 & 0.6200 \\
			GFP-GAN \cite{wang2021towards}     & \textcolor{blue}{0.3646} & \textcolor{blue}{42.62}  & \textcolor{red}{4.077}  & 25.08 & 0.6777 \\ \hline
			\textbf{FaceFormer(ours)} & \textcolor{red}{0.3237}  & \textcolor{red}{13.17} & \textcolor{blue}{4.369} &  \textcolor{red}{26.94}  &   \textcolor{red}{0.7386}     \\ \hline  
		\end{tabular}
	\end{center}
\end{table}
\setlength{\tabcolsep}{1.4pt}

For training, the proposed FaceFormer is conduct with Adam optimizer of total 800k iterations. We use a learning rate of $2e^{-3}$ with a MultiStepLR decay policy by a factor of 2 at the 700k-th, 750k-th iterations. We implement our models with the PyTorch framework and train them using two NVIDIA Tesla A100 GPUs. 

\begin{figure}[t]
	\centering
	\includegraphics[width=12cm]{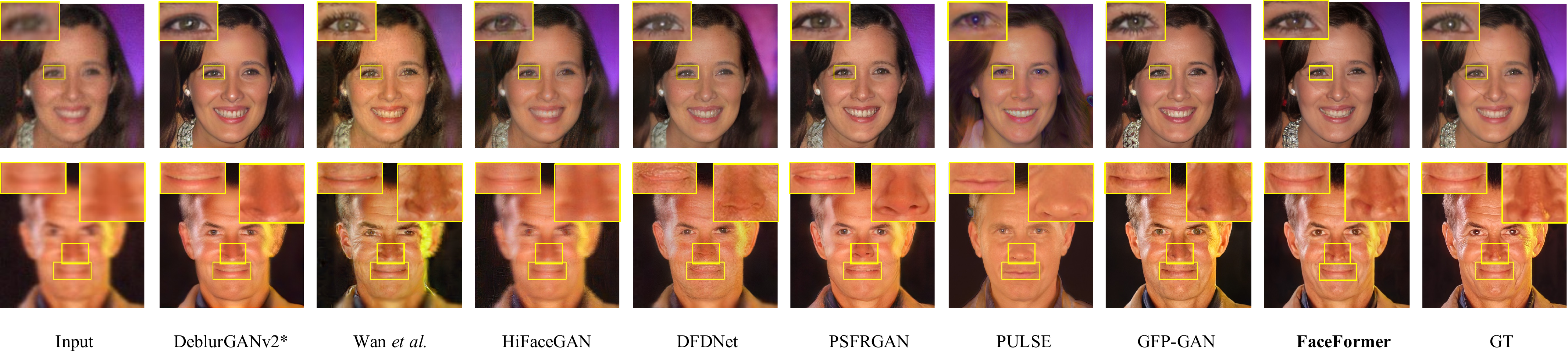}
	\caption{Qualitative comparison on the CelebA-Test for blind face restoration. Our GFP-GAN produces faithful details in eyes, mouth and hair. Zoom in for best view.}
	\label{fig:syth_data_comparisons}
\end{figure}

\subsection{Comparisons with State-of-the-art Methods}
We compare the proposed FaceFormer with some state-of-the-art methods designed for different restoration tasks: HiFaceGAN \cite{yang2020hifacegan}, DFDNet \cite{li2020blind}, PSFRGAN \cite{chen2021progressive}, Super-FAN \cite{bulat2018super}, Wan et al. \cite{wan2020bringing}, GFP-GAN \cite{wang2021towards}, PULSE  \cite{menon2020pulse} and mGANprior \cite{gu2020image} for face restoration; DeblurGANv2 \cite{kupyn2019deblurgan} for natural image super-resolution. 
Specially, HiFaceGAN, DFDNet and GFP-GAN are conducted for blind face restoration. Note that the above methods are adopted their official codes except for Super-FAN, for which we use a re-implementation.


\setlength{\tabcolsep}{4pt}
\begin{table}
	\begin{center}
\caption{Quantitative comparison on the real-life LFW, CelebChild, WebPhoto. \textcolor{red}{Red} and \textcolor{blue}{blue} indicates the best and the second best performance. $(*)$ denotes finetuning on our training set.}
\label{table:real_world_comparisons}
	\begin{tabular}{l|ll|ll|ll}
		\hline
		Dataset          & \multicolumn{2}{l|}{LFW-Test} & \multicolumn{2}{l|}{CelebChild} & \multicolumn{2}{l}{WebPhoto} \\
		Methods          & FID $\downarrow$ & NIQE $\downarrow$ & FID $\downarrow$ & NIQE $\downarrow$ & FID $\downarrow$ & NIQE $\downarrow$ \\ \hline
		Input            & 137.56         & 11.214        & 144.42          & 9.170         & 170.11        & 12.755       \\ 
		DeblurGANv2$(*)$  \cite{kupyn2019deblurgan}    & 57.28          & 4.309         & 110.51          & 4.453         & 100.58        & 4.666        \\ 
		Wan et al. \cite{wan2020bringing}       & 73.19          & 5.034         & 115.70          & 4.849         & 100.40        & 5.705        \\
		HiFaceGAN \cite{yang2020hifacegan}         & 64.50          & 4.510         & 113.00          & 4.855         & 116.12        & 4.885        \\
		DFDNet \cite{li2020blind}            & 62.57          &  \textcolor{blue}{4.026}  & 111.55          & 4.414         & 100.68        & 5.293        \\
		PSFRGAN \cite{chen2021progressive}          & 51.89          & 5.096         & 107.40          & 4.804         & 88.45         & 5.582        \\ \hline
		mGANprior \cite{gu2020image}         & 73.00          & 6.051         & 126.54          & 6.841         & 120.75        & 7.226        \\
		PULSE \cite{menon2020pulse}            & 64.86          & 5.097         & \textcolor{blue}{102.74}   & 5.225         & \textcolor{red}{86.45}   & 5.146        \\ 
		GFP-GAN  \cite{wang2021towards}          & \textcolor{blue}{49.96}  & \textcolor{red}{3.882} & 111.78          & \textcolor{blue}{4.349}   & \textcolor{blue}{87.35} & \textcolor{blue}{4.144}     \\ \hline
		\textbf{FaceFormer(ours)} & \textcolor{red}{49.36}   & 4.166   &   \textcolor{red}{72.27}  &   \textcolor{red}{4.293}   &  137.59  &   \textcolor{red}{4.1341}   \\ \hline
	\end{tabular}
	  \end{center}
\end{table}
\setlength{\tabcolsep}{1.4pt}

For the evaluation, we employ the widely-used non-reference perceptual metrics: FID \cite{heusel2017gans} and NIQE \cite{mittal2012making} for real-life low quality faces without Ground-Truth (GT). These metrics measure the realism and statistic distance between the restorations and inputs. Smaller values indicate better restorations. In addition, we also adopt pixel-wise metrics (PSNR and SSIM) and the perceptual metric (LPIPS \cite{zhang2018unreasonable}) for the CelebA-Test with GT. The LPIPS evaluates the similarity of deep features between restorations and GTs to simulate human perceptual similarity. Smaller values indicate closer similarity.

\subsubsection{Synthetic CelebA-Test} The comparisons are conducted under these setting that blind face restoration whose inputs and outputs have the same resolution. 
Table~\ref{table:celeba_comparisons} reports the quantitative results among comparison methods. It is can be observed that our FaceFormer get the lowest LPIPS score, indicating that FaceFormer is perceptually close to GT. Our FaceFormer also surpasses other methods by a large margin in terms of FID, and is 29$\%$ higher than the second best result of GFPGAN. It shows that the restoration is close to the real face distribution in statistical similarity of Inception \cite{szegedy2017inception} feature embedding. Besides the perceptual performence, our method also reports the  pixel-wise metrics PSNR and SSIM. It owns to the proposed Facial Feature Embedding, which learns detailed geometrical structure features.  

Qualitative results are presented in Fig.~\ref{fig:syth_data_comparisons}. Thanks to the robustness FFE module, our FaceFormer recovers faithful details in the eyes (pupils and eye-lashes), nose, etc. Our FaceFormer is capable of retaining fidelity, e.g., it produces naturally closed mouth without forced addition of teeth as PSFRGAN does. Moreover, the mouth generated by FaceFormer perceptually looks more similar to GT than GFP-GAN.


\begin{figure}[t]
	\centering
	\includegraphics[width=12cm]{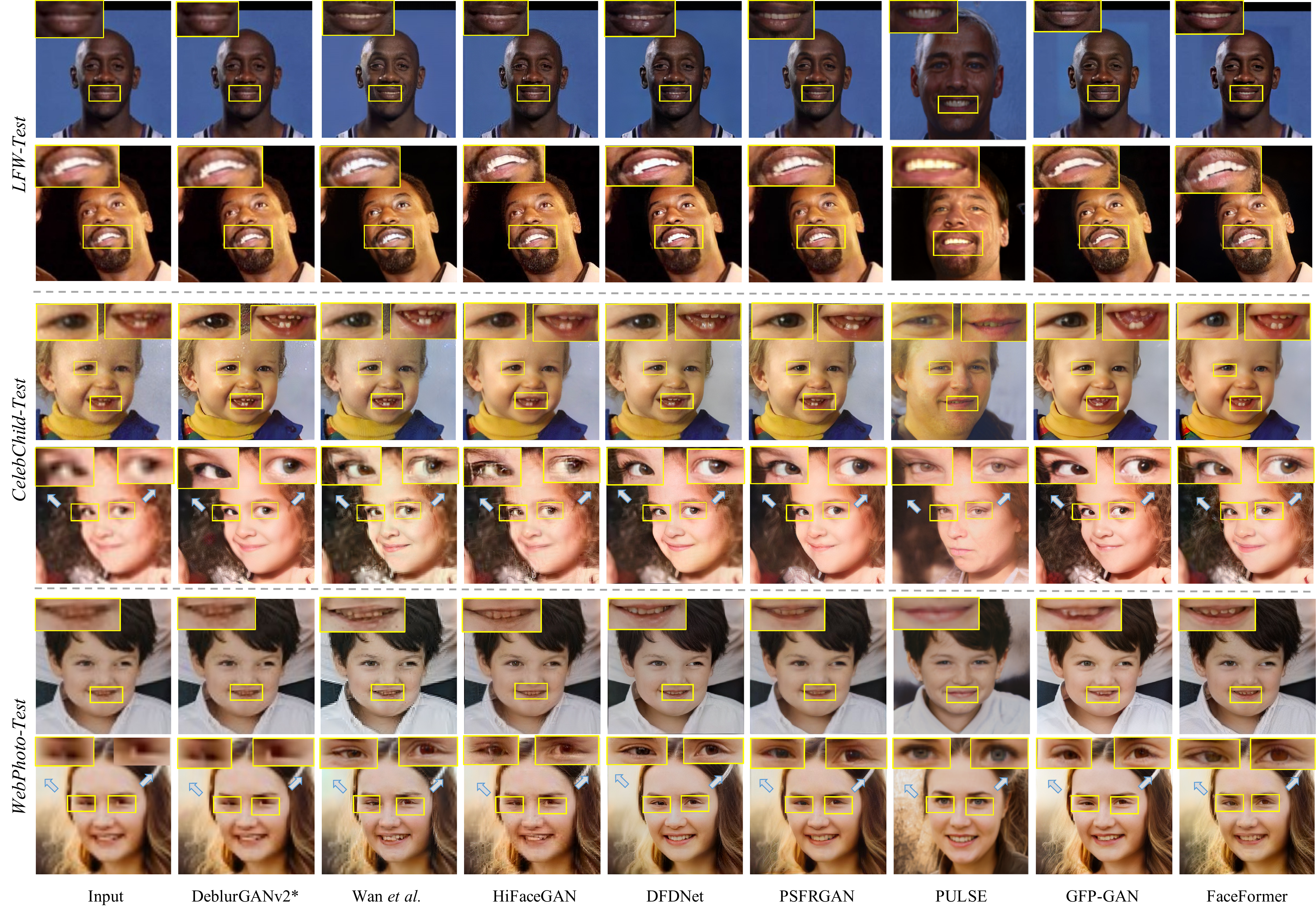}
	\caption{Qualitative comparisons on three real-life datasets. Zoom in for best view.}
	\label{fig:real_world_comparisons}
\end{figure}

\begin{figure}[b]
	\centering
	\includegraphics[width=10cm, height=3cm]{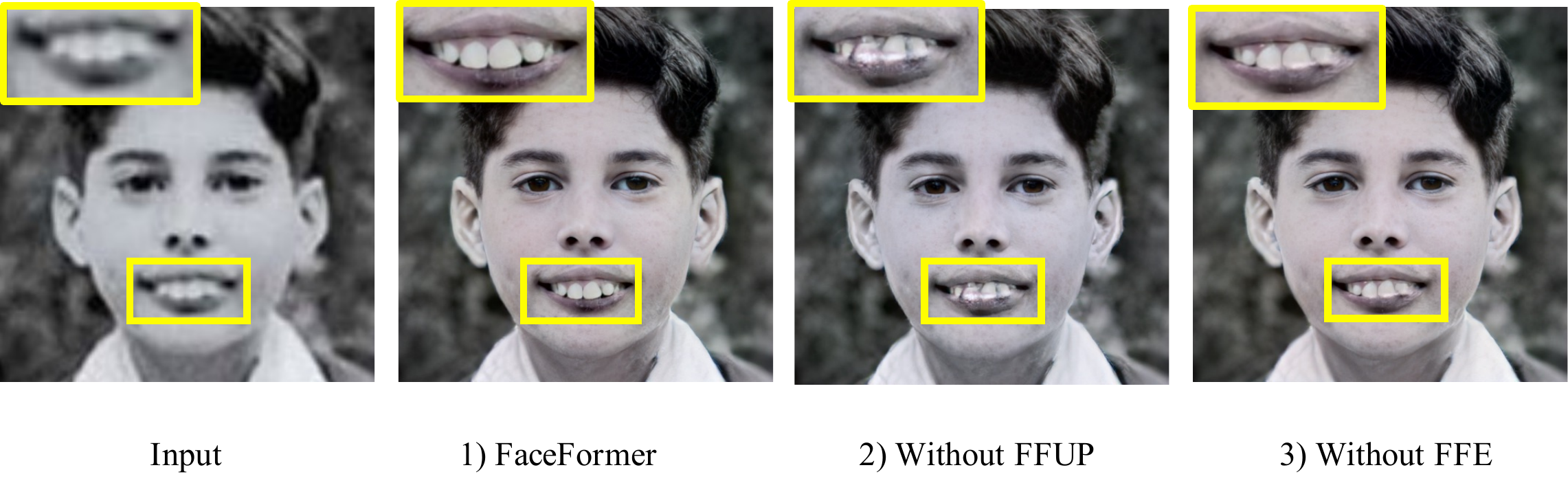}
	\caption{Ablation studies on the effectiveness of FFUP module, FFE module. Zoom in for best view.}
	\label{fig:albation_1}
\end{figure}

\subsubsection{Real-Life LFW, CelebChild and WedPhoto-Test}  The final target of all methods is to restore real world LQ face images. To evaluate the generalization ability of comparison methods, we also evaluate the performance of FaceFormer on three different real-world datasets. The quantitative results are shown in Table~\ref{table:real_world_comparisons}. Our method achieves surpassing performance on LFW-test and CelebChild datasets, showing its remarkable generalization capability. 

Fig.~\ref{fig:real_world_comparisons} reports the qualitative comparisons overall comparison methods. Our method could produce plausible and realistic faces on complicated real-world degradation while other methods fail to recover faithful facial details or produce transparent artifacts (especially in CelebChild-Test in Fig ~\ref{fig:real_world_comparisons}). 

Our method takes arbitrary scale face as input and generates scale-aware realistic restorations through FFUP module. Thus, the restorations largely avoid the transparent artifacts caused by scale drift, especially at the boundary of face components. In addation, FaceFormer keep the same pupil color in the 4th row owing to FFE module.

It can be observed that results of HiFaceGAN, Wan et al. and GFPGAN contains artifacts in the mouth and eyes. This is most likely because of the unstable training of GAN without face prior guidance. Different from PSFR-GAN which utilizes parsing map, DFDNet needs to first detect the LQ face components and then matches them to a HQ dictionary. It may fail to find the correct reference when the LQ are too blurry or with large pose. For example, the results of DFDNet in the 3-nd row seems to generate blur teethes.

\setlength{\tabcolsep}{4pt}
\begin{table}
	\begin{center}
		\caption{Ablation study results on CelebA-Test under blind face restoration.}
		\label{table:ablation_comparisons}
		\begin{tabular}{l|lll|ll}
			\hline
			Configuration                             & LPIPS $\downarrow$ & FID $\downarrow$ & NIQE $\downarrow$ & PSNR $\uparrow$ & SSIM $\uparrow$ \\ \hline
			1) Our FaceFormer with FFUP  & 0.3237   & 13.17        & 4.369       &  26.94     &   0.7386    \\ \hline
			2) Our FaceFormer w/o FFUP   & 0.3502 ($\uparrow$) & 17.82 ($\uparrow$)  & 4.481 ($\uparrow$) &  25.36 ($\downarrow$)    &  0.7182 ($\downarrow$)  \\ \hline
			3) Our FaceFormer w/o FFE   & 0.3646 ($\uparrow$) & 42.62 ($\uparrow$)   & 4.077 ($\downarrow$)      & 25.08  ($\downarrow$)   &  0.6777  ($\downarrow$)   \\ \hline
		\end{tabular}
	\end{center}
\end{table}
\setlength{\tabcolsep}{1.4pt}

\subsection{Ablation Studies}

For ablation study, we test the proposed FaceFormer on CelebA-test set to discuss the impact of each module.

\subsubsection{Impact of the FFUP.} 

As shown in Table.~\ref{table:ablation_comparisons} [configuration 2)] and Fig.~\ref{fig:albation_1}, when we remove the Facial Feature Up-sampling layer, it can be observed that the restored faces produce apparent artifacts, especially in the teeth. The appearance of image artifacts is usually caused by the reduction of high-frequency spatial frequency, which is usually accompanied by the change of object edge width, mainly in the process of image re-sampling. Thus, the face identity cannot be well retained without the FFUP module (high perceptual metrics). 

\subsubsection{Robustness to multi-scale Face Inputs.}  
\begin{figure}[t]
	\centering
	\includegraphics[width=\linewidth, height=6cm]{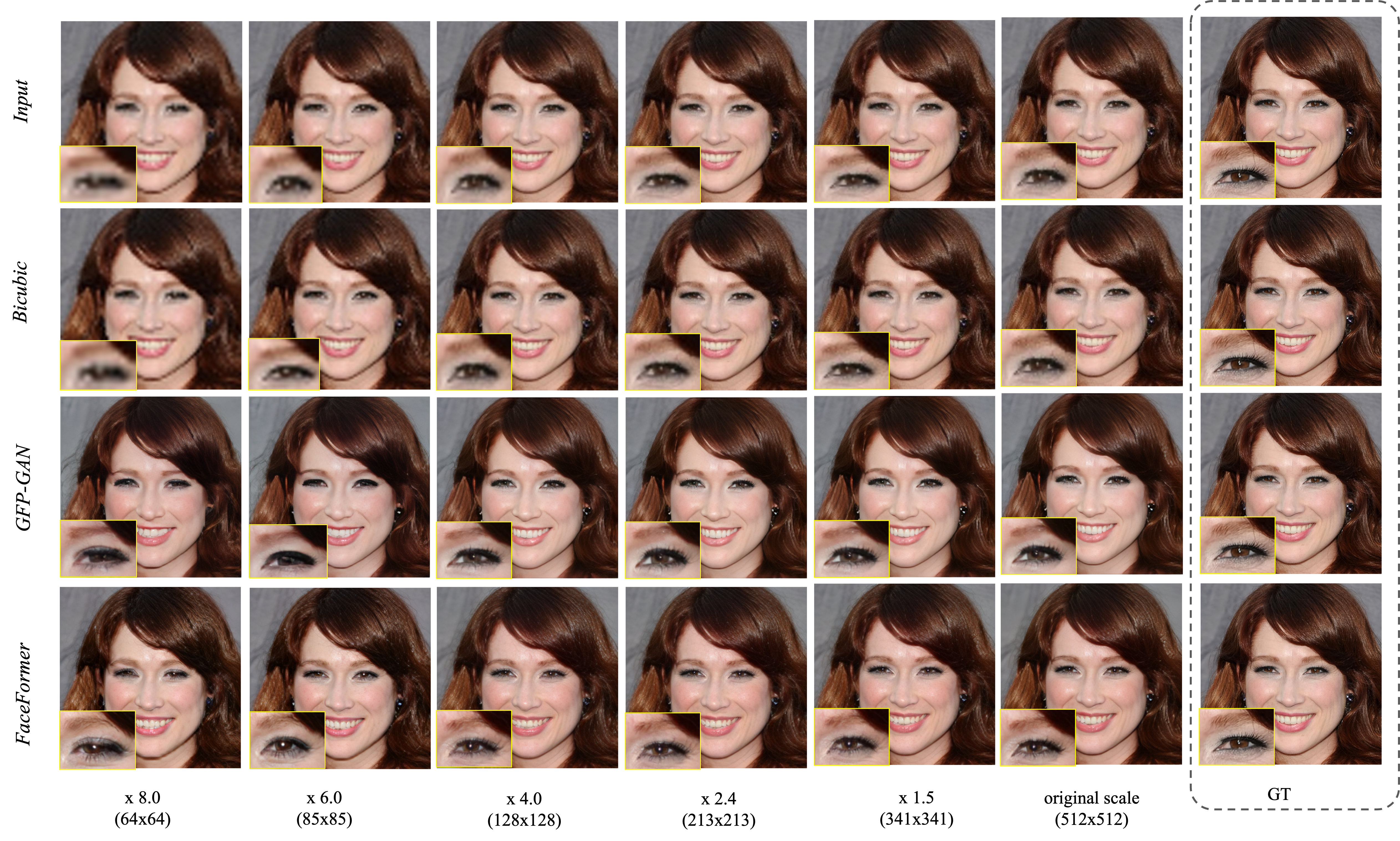}
	\caption{Results for different input resolutions while fixing other degradation parameters. The `($\cdot$)' denotes the input resolution.}
	\label{fig:multi_scale}
\end{figure}

We conduct comparions of the arbitrary scale inputs. Specifically, the scale of $\left\{ \times 1.5, \times 2.4, \times 4.0, \times 6.0, \times 8.0 \right\}$ are adopt to the FFUP module as the upsampling scale, respectively. As shown in Fig.~\ref{fig:multi_scale}, our FaceFormer works well for upscale factors and can produce more reasonable results, such as more realistic eyebrows and skin texture. Note that when the upscale factor is large ($\times 6.0$, $\times 8.0$ scale), the FaceFormer restorations are perceptually superior to the GFP-GAN ones.
We further provide additional visualizations of diverse upscale factors in supplemental material.

\subsubsection{Impact of the FFE module.} The FFE module is designed to provide diversity and robustness of facial latent feature, and strengthen the restoration ability for complicated degradation in the real world. Without the FFE module, as shown in Table.~\ref{table:ablation_comparisons} [configuration 3)] and Fig.~\ref{fig:albation_1}, it can be observed that the facial components are incomplete for the lack of local shape information, especially in teeth. Hence, the generated face has suboptimal fidelity in the absence of FFE module (high perceptual metrics and low pixel-wise metrics).


\section{Conclusions}

This paper proposes a scale-aware blind face restoration framewark FaceFormer. The FaceFormer formulates facial feature restoration as scale-aware transformation procedure and then adopts hierarchical transformer embedding blocks to achieve high fidelity, realistic and symmetrical details of facial components with arbitrary scale inputs. Extensive comparisons demonstrate the superior capability of FaceFormer in joint face restoration for real-world images, outperforming prior art.

\bibliographystyle{splncs}
\bibliography{egbib}

\end{document}